\pdfoutput=1

\documentclass[11pt]{article}

\usepackage{acl}
\usepackage{times}
\usepackage{latexsym}

\usepackage[T1]{fontenc}

\usepackage[utf8]{inputenc}

\usepackage{microtype}
\usepackage{mathtools}
\usepackage{amsmath}
\usepackage{amsfonts}
\usepackage[ruled,noline]{algorithm2e}
\usepackage[final]{graphicx}
\usepackage{subcaption}
\usepackage{verbatim}

\usepackage{amsmath}
\DeclareMathOperator*{\argmax}{arg\,max}

\newcommand{\modelname}{\texttt{DPAV}}

\title{Anti-Overestimation Dialogue Policy Learning for Task-Completion Dialogue System}

\author{Chang Tian$^{\dagger}$ \and Wenpeng Yin$^*$ \and Marie-Francine Moens$^{\dagger}$\\
  $^{\dagger}$Department of Computer Science, KU Leuven\\
  $^*$LanguageX Lab, Temple University\\
  \texttt{chang.tian@kuleuven.be}}

\begin{document}
\maketitle
\begin{abstract}
A dialogue policy module is an essential part of task-completion dialogue systems. Recently, increasing interest has focused on reinforcement learning (RL)-based dialogue policy. Its favorable performance and wise action decisions rely on an accurate estimation of action values. The overestimation problem is a widely known issue of 
RL since its estimate of the maximum action value is larger than the ground truth, which results in an unstable learning process and suboptimal policy. This problem is detrimental to RL-based dialogue policy learning.
To mitigate this problem, this paper proposes a dynamic partial average  estimator (\modelname) of the ground truth maximum action value. \modelname~calculates the partial average between the predicted \textit{maximum} action value and \textit{minimum} action value, where the weights are dynamically adaptive and problem-dependent. We incorporate \modelname~into a deep Q-network as the dialogue policy and. Our method can achieve better or comparable results compared to top baselines on three dialogue datasets of different domains \textit{with a lower computational load}. In addition, we also theoretically prove the convergence and derive the upper and lower bounds of the bias compared with those of other methods.
\end{abstract}

\section{Introduction}
\begin{figure*}[ht!]
\begin{subfigure}[b]{.5\textwidth}
  \centering
  \includegraphics[width=0.9\textwidth]{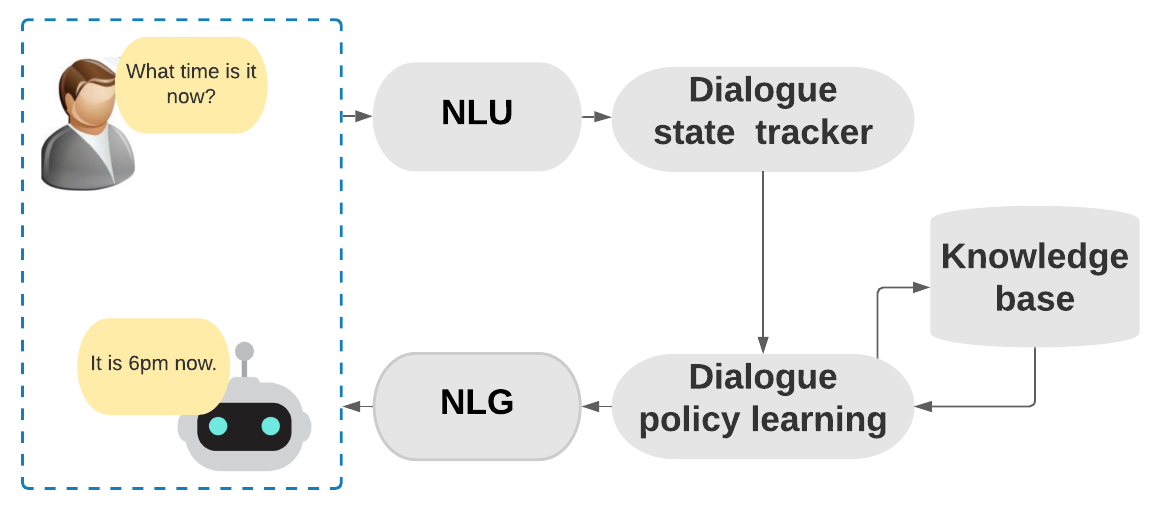}
  \caption{Four modules in the pipeline framework}
  \label{img:TDS}
\end{subfigure}%
\begin{subfigure}[b]{.5\textwidth}
  \centering
  \includegraphics[width=0.65\textwidth]{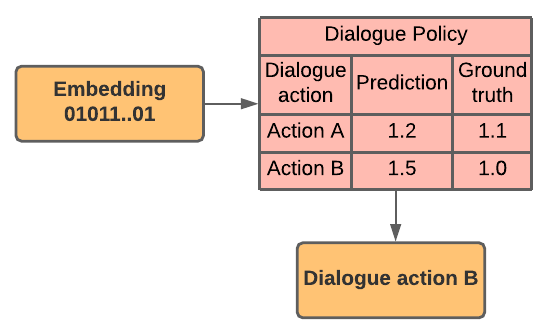}
  \caption{The overestimation problem in RL-based dialogue policy}
  \label{img:pd}
\end{subfigure}
\caption{Task-completion dialogue system}
\label{fig:dialoguepipeline}
\end{figure*}

Task-completion dialogue systems are commonly implemented in two schemes. One is by end-to-end training, such as \cite{zhang2020probabilistic}. The other is a pipeline framework~\citep{chen2017survey}, 
which typically consists of four modules that are independently trained, as shown in Figure \ref{img:TDS}: natural language understanding (NLU), dialogue state tracker (DST), dialogue policy learning (DPL) and natural language generation (NLG). For this pipeline-style dialogue system,  the conversation text from a user is first fed to the NLU module, where the user utterance is parsed into semantic slots for DST. DST manages the inputs of each dialogue turn together with the dialogue history. Then DST outputs the current dialogue state embedding to the DPL module, where a dialogue action is taken based on current dialogue state and knowledge base data. The NLG module maps the selected dialogue action into natural language to converse with the user.

\begin{figure}[ht!]
\includegraphics[width=0.477\textwidth]{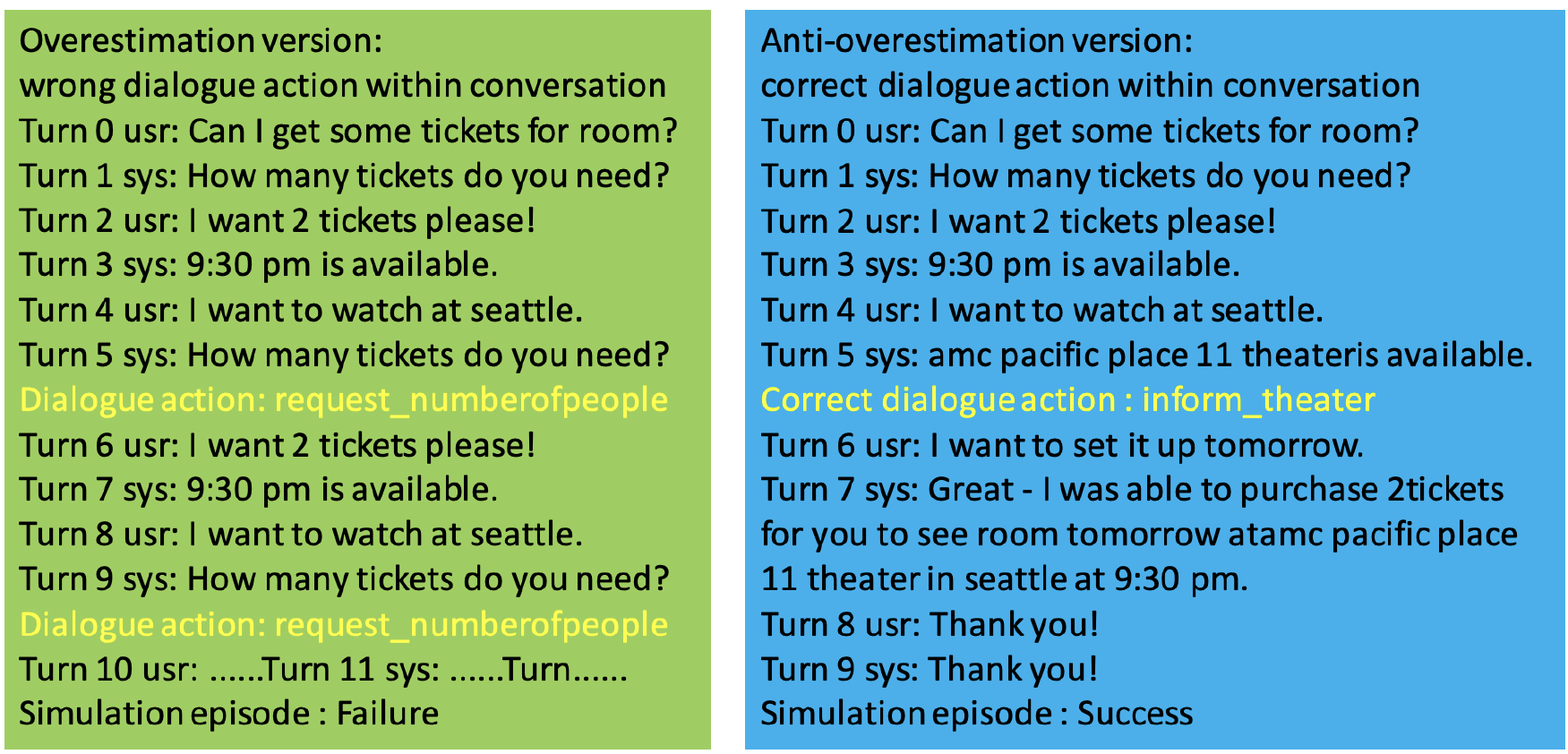}
\centering
\caption{Wrong and correct dialogue actions}
\label{img:dialogue_example}
\end{figure}

Reinforcement learning (RL) algorithms, specifically 
Q-learning~\citep{watkins1992q} based algorithms, have become a mainstream method for training the dialogue policy module~\citep{peng2018deep, zhang2020recent}. For each step, the policy agent updates its action value \footnote{This value is the expected return for taking the action under a certain state, and it is represented as the Q value of Q-learning.} estimate as the sum of the observed reward and the estimated maximal action value in the next state. However, this update rule suffers from an overestimation problem~\citep{hasselt2010double}: mostly the estimated maximal action value is larger than the ground truth. The overestimation problem causes that the dialogue policy module has inaccurate action values estimations after the training, which misleads the dialogue policy to choose the wrong dialogue action (see the wrong dialogue action in Figure~\ref{img:dialogue_example}). Some prior studies have tried to address this problem in domains like video game playing and multi-agent systems, but they either suffered from the underestimation problem \citep{hasselt2010double,lan2020maxmin} or required heavy computational load, such as those ensemble methods \citep{anschel2017averaged,lan2020maxmin,lee2021sunrise}.

In this work, we propose dynamic partial average (\modelname), a novel approach to mitigate the overestimation problem specifically for the task-completion dialogue policy. \modelname~utilizes the \textit{partial average} between the predicted \textit{maximal} action value and the predicted \textit{minimal} action value to estimate the ground truth maximum action value, where the weights are dynamically adaptive and problem-dependent. The rationale here is that \modelname~ learns the optimal trade-off between the predicted maximal action value and the predicted minimal action value so that the dialogue policy learning procedure will be more reasonable and stable. Our system not only yields a better dialogue process (see Figure~\ref{img:dialogue_example}), but also has much lower computational cost compared to ensemble models. 

Overall~\footnote{The project resources are in the GitHub \url{https://github.com/changtianluckyforever/version_one}.}, our main contributions are as follows: (i) This is the first work to investigate and handle the overestimation problem of the reinforcement learning framework for task-completion dialogue systems. (ii) We propose a novel and effective approach, the dynamic partial average \modelname, which can alleviate the overestimation problem with lower computational load. (iii) We theoretically prove the convergence and derive the upper and lower bounds of our method to claim its effectiveness. 
\section{Related Work}
\paragraph{Dialogue Policy.} The dialogue policy module makes a dialogue decision given the current state~\citep{zhang2019budgeted}. Early methods are rule-based \citep{chen2017survey}. Since handcrafted rules are non-extensible and resource-consuming~\citep{zhao2021efficient}, deep reinforcement learning (DRL) has become a mainstream method for training dialogue policies~\citep{wu2019switch,wang2020task, zhao2021efficient}. Task-completion dialogue policy learning is often regarded as an RL problem~\citep{zhang2021emotion}.

\paragraph{Overestimation Bias.} 
The value-based algorithm Q-learning, a common unit of the dialogue policy module, suffers the overestimation bias~\citep{thrun1993issues, hasselt2010double}. Prior studies addressed the problem in multiple ways, including (1) bias compensation with additive pseudo costs and (2) a variety of estimators. Bias-corrected Q-Learning~\citep{lee2013bias} subtracts a quantity from the target but this method cannot address the bias from the function approximation~\citep{pentaliotis2021variation}. It is known that the bias compensation method is 
labor involved and time consuming~\citep{anwar2008actor,lee2012intelligent}. Double Q-learning~\citep{hasselt2010double} trades overestimation bias for underestimation bias using the double estimator. Since underestimation bias is not preferable~\citep{hasselt2010double,lan2020maxmin},  Weighted Q-learning proposes~\citep{d2016estimating,zhang2017weighted} the weighted estimator for the maximal action value based on a weighted average of estimated actions values. However, the weights computation is only practical in a tabular setting~\citep{d2017estimating}. Our work differs from the foregoing in that it proposes a new estimator which could be generalized into the deep Q-learning network setting. 

Overestimation bias is more problematic in the deep Q-learning network (DQN) algorithm~\citep{fan2020theoretical} due to the function approximation errors of DRL. Polishing estimation tricks of a single model and using ensemble models are two mainstream solutions. Double Q-learning is subsequently adapted to a neural network as Double DQN~\citep{van2016deep}, and Duel DQN proposes a new action value estimation scheme~\citep{wang2016dueling}. But the two methods still suffer the bias of double estimator and maximum estimator, respectively. Another approach against overestimation bias is based on the idea of ensembling. Averaged DQN controls the estimation bias by taking the average over action values of multiple target networks~\citep{anschel2017averaged}. Later, \citep{lan2020maxmin} claims that an average operation will never completely remove the overestimation bias, and they propose the Maxmin DQN which takes a minimum from multiple maximums of different ensemble units to estimate the maximum action value in a selective process. Then, \citep{kuznetsov2020controlling} recognizes that Maxmin DQN also suffers underestimation bias and that the bias control is coarse. Recently, the SUNRISE method uses the uncertainty estimates of the ensemble. But it only down-weights the biased estimation~\citep{lee2021sunrise}. In this work, the model only uses a value function instead of a combination of multiple value functions and tailors the predicted maximum and minimum of a value function to approximate the optimal action value. Our work does not move towards underestimation and avoid the computational complexity of ensemble models.
\section{Preliminary}
\subsection{Problem Definition}
Even though an unbiased estimator does not exist~\citep{d2016estimating}, the maximum estimator~\citep{watkins1992q} and double estimator~\citep{hasselt2010double, van2016deep}  are the most representative among the relevant works.
\paragraph{Maximum estimator (ME).} 
This method is used by deep Q-learning to approximate the ground truth maximum action value of the following state by maximizing over a set of  action values  $Q\left(s_{t+1}, \cdot\right)$. It represents the target $y^{D Q N}$ for taking a possible action  $a$ under the state $s_{t+1}$ as: 
\begin{align}
y^{D Q N}=r_{t+1}+\gamma \max _{a} Q\left(s_{t+1}, a ; \boldsymbol{\theta}^{-}\right)
\end{align}
where $r_{t+1}$ is the reward, $\gamma$ is the discount value for future rewards and $\boldsymbol{\theta}^{-}$ is parameters of the target network.  
As \newcite{smith2006optimizer} found, the estimate of ME is larger than the ground truth (i.e., the estimated maximum value of the following state, $\max _{a} Q\left(s_{t+1}, a ; \boldsymbol{\theta}^{-}\right)$ is overestimated), which results in the biased loss:
\begin{equation}
\begin{aligned}
&L(\boldsymbol{\theta})=\\&\underset{\left\langle s_{t}, a_{t}, r_{t}, s_{t+1}\right\rangle \sim m}{\mathbb{E}}\left[\left(y^{D Q N} -Q\left(s_{t}, a_{t} ; \boldsymbol{\theta}\right)\right)^{2}\right],
\end{aligned}
\label{eq:dqnloss}
\end{equation}
where $m$ is the RL experience replay pool and $\boldsymbol{\theta}$ is parameters of the DQN model. Thus, the $Q\left(s_{t}, \cdot\right)$ will not be perfectly accurate after training, 

\paragraph{Double estimator (DE).} 
This method~\citep{hasselt2010double,van2016deep} 
is used by deep Q-learning to solve the overestimation problem of ME in DQN. The Double DQN has two estimators, and one estimator decides the action index while the other estimator evaluates the action value of the selected action. Then Double DQN (DDQN) uses the evaluated action value to estimate the ground truth maximum action value of state $s_{t+1}$:
\begin{equation}
\begin{aligned}
&y^{D D Q N}=\\&r_{t+1}+\gamma Q(s_{t+1}, \underset{a}{\operatorname{argmax}} Q\left(s_{t+1}, a ; \boldsymbol{\theta^{+}}\right) ; \boldsymbol{\theta}^{-}).
\end{aligned}
\end{equation}
However, DE suffers from the underestimation problem and does not guarantee better estimation than ME~\citep{lan2020maxmin}.
\subsection{Problem in Dialogue Policy}
Q-learning is a common unit of RL-based dialogue policies. The overestimation bias of ME propagates into model action values $Q\left(s_{t}, \cdot\right)$. In dialogue $Q\left(s_{t}, \cdot\right)$ represent the dialogue action values, which are the expected returns the dialogue system will receive after taking an action under the state $s_{t}$. Since $Q\left(s_{t}, \cdot\right)$ are biased, the dialogue policy cannot issue accurate actions accordingly. This hurts dialogue performances.

\paragraph{Example.} We use a dialogue turn to show the negative effects of the overestimation bias. In Figure~\ref{img:pd}, the dialog state tracker module outputs state embedding, dialogue policy processes state embedding and predicts the wrong dialogue action B instead of the correct action A based on the biased action values. 
\section{Method}
\subsection{Dynamic Partial Average}
Q-learning suffers from overestimation bias because of the ME~\citep{hasselt2010double}. To reduce the bias, in this work, we propose the dynamic partial average (DPAV) estimator. DPAV utilizes the partial average between the predicted maximal action value and the minimal action value to estimate the ground truth maximal action value $Q_{*}(s_{t+1})$ of the target of Q-learning update,
The mathematical formula of the DPAV estimator of Q-learning is as follows:
\begin{equation}~\label{eq:optimal}
\begin{aligned}
&Q_{*}(s_{t+1})\approx (1-\lambda_{t})*\max _{a^{\prime}} Q\left(s_{t+1}, a^{\prime} ; \theta^{\prime}\right)\\&+ \lambda_{t}*\min _{a^{\prime \prime}} Q\left(s_{t+1}, a^{\prime \prime} ; \theta^{\prime}\right)=Q_{DPAV}(s_{t+1}),
\end{aligned}
\end{equation}
$\lambda_{t}$ is a float number between [0,1] that is dynamic in time and problem-dependent such that the DPAV can take the average between the maximum and minimum of the action values. The weights assigned to the maximum and minimum are not the same, so it is a partial average.

The DPAV estimator is deployed in Q-learning as DPAV Q-learning, so that we have the action value function Q update formula as:
\begin{equation}~\label{eq:1}
\begin{aligned}
&Q\left(s_{t}, a_{t}\right) \leftarrow (1-\alpha_{t}) Q\left(s_{t}, a_{t}\right)+\alpha_{t} y^{DPAV}, \\
&y^{DPAV}= r_{t+1}+\gamma*Q_{DPAV}(s_{t+1}). 
\end{aligned}
\end{equation}
where $\gamma$ is the discount factor for the future action value and $\alpha_{t}$ is the step size. $\lambda_{t}$ in $Q_{DPAV}(s_{t+1})$ decays according to a predefined rate as training progresses, the decay formula is as follows:
\begin{equation}
\lambda_{t+1}= \lambda_{t} * d,
\end{equation}
where $d$ is the decay rate that is set to the fixed value in training. So $1-\lambda_{t}$ will give more weight to the maximal action value during the training.

To apply DPAV to the complex dialogue policy learning setting, this paper combines it 
with the deep Q-learning network (DQN) and proposes DPAV DQN. 
Its loss function is adapted from Equation~\ref{eq:dqnloss} to the formula:
\begin{equation}
\begin{aligned}
&L_\theta=\\&\underset{\left\langle s_{t}, a_{t}, r_{t+1}, s_{t+1}\right\rangle \sim m}{\mathbb{E}}\left[\left(y^{DPAV} -Q\left(s_{t}, a_{t} ; \theta\right)\right)^{2}\right].
\end{aligned}
\end{equation}
The algorithm of the dynamic partial average deep Q-learning network is summarized in Algorithm~\ref{alg:dpav}.\footnote{In the algorithm, if the state $s_{t+1}$ is a terminal state, it means the Markov decision process ends. And in the dialogue, it means the dialogue ends.} 

The intuition behind this approach is that the predicted maximal action value overestimates the ground truth, so DPAV uses the predicted minimal action value to shift the estimate towards the ground truth. Because the predicted action values accuracy is improved in training, DPAV assigns less weight to the predicted minimum to avoid shifting towards the small estimate too much.
DPAV reduces the overestimation bias in the target of the training loss, so it is less biased. This improves the dialogue action values accuracy of the DPAV DQN dialogue policy, so this dialogue policy issues more accurate dialogue actions accordingly which improve dialogue performances.

Additionally, this method has a lower computational complexity compared to those of ensemble models. Even if the latter could trade time complexity for space complexity by parallel computing, they still have high computational complexity in general as shown in the Table~\ref{tab:comple}. And this method achieves better or comparable performances according to the Figure~\ref{fig:all_tests}. The upper and lower bounds of the DPAV DQN estimation bias are also reasonable compared with those of other methods. A detailed explanation is found in section 4.3.

\begin{algorithm}[hbt!]
\SetAlgoLined
Initialize replay memory \(\mathcal{D}\) to capacity \(N\), action-value function \(Q\) with random weights, and decay rate $d$ \\
 \For{episode =1,...,M}{
 Initialise state \(s_{1}\)\\
 \For{j=1,...,T}{
    With probability \(\epsilon\) select a random action \(a_{j}\), otherwise select \(a_{j}=\max _{a} Q^{*}\left(s_{j}, a ; \theta\right)\)\\
    Execute action \(a_{j}\) in environment, observe reward \(r_{j+1}\) and come into state \(s_{j+1}\). Store transition \(\left(s_{j}, a_{j}, r_{j+1}, s_{j+1}\right)\) in \(\mathcal{D}\), and set \(s_{j}=s_{j+1}\)\\
    Sample random minibatch of \(\left(s_{t}, a_{t}, r_{t+1}, s_{t+1}\right)\) from \(\mathcal{D}\)
  $$\text { Set } y_{t}=\left\{\begin{array}{l}
{r_{t+1}}\\ {\text {if terminal state } s_{t+1}}\\
{y^{DPAV}}\\ {\text {non-terminal state } s_{t+1}}
\end{array}\right.$$
Perform a gradient descent step on $L_\theta$=\(\left(y_{t}-Q\left(s_{t}, a_{t} ; \theta\right)\right)^{2}\) to update $\theta$\\
Replace target parameter $\theta^{-} \longleftarrow \theta$ after every \textit{L} iterations. Update average weight $\lambda_{t+1} \longleftarrow \lambda_{t}*d $ after every \textit{U} iterations.
}
}
\caption{DPAV DQN} 
\label{alg:dpav}
\end{algorithm}

\subsection{Convergence}
In this subsection we show in Theorem 1 that in the limit DPAV Q-learning converges to the optimal policy. The proof\footnote{Lemma 1 was also used to prove the convergence of SARSA~\citep{rummery1994line} and Double Q-learning~\citep{van2016deep} } of this result using the Lemma 1~\citep{singh2000convergence} is in the Appendix~\ref{sec:appendixdpavpr}.

\paragraph{Theorem 1.} In a Markov decision process, the approximate action value function $Q$ as updated by DPAV Q-learning in Equation~\ref{eq:1} converges to the optimal action value function $q_{*}$ with probability one if an infinite number of experience tuples in the form of $\left(s_{t}, a_{t}, r_{t+1}, s_{t+1}\right)$ are given by a learning policy for each state action pair and if the following conditions are satisfied:\\
1. The Markov decision process is finite (i.e. $\mid \mathcal{S} \times$ $\mathcal{A} \times \mathcal{R} \mid<\infty$, $\mathcal{S}$ means the set of states, $\mathcal{A}$ means the set of actions, and $\mathcal{R}$ is the set of rewards.).\\
2. $\gamma \in[0,1)$.\\
3. $\alpha_{t}(s, a) \in[0,1]$, $\sum_{t} \alpha_{t}(s, a)=\infty, \sum_{t}\alpha_{t}^{2}(s, a)<\infty$ $w . p .1$, and $\forall s, a \neq s_{t}, a_{t}: \alpha_{t}(s, a)=0$. $\alpha_{t}(s, a)$ is the step size of a Q-learning update.
\subsection{Upper and Lower Bound}
As shown in~\citep{d2016estimating, hasselt2010double}, considering a set of $M \geq 2$ independent random variables $X=\left\{X_{1}, \ldots, X_{M}\right\}$, each random variable $X_{i}$ has a mean $\mu_{i}=\mathbb{E}\left[X_{i}\right]$ and a variance  $\sigma_{i}=\operatorname{Var}\left[X_{i}\right]$. In many problems, one is interested in the maximum expected value in such a set $\mu_{*}=\max _{i} E\left\{X_{i}\right\}$. Without knowledge of the functional form and parameters of the underlying distribution of each variable $X_{i}$, it is impossible to find $\mu_{*}$ analytically. Given a set of a limited number of samples,
$S=\left\{S_{1}, \ldots, S_{M}\right\}$, $S_{i}$ corresponds to the subset of samples drawn from the unknown distribution of the random variable $X_{i}$. The maximum estimator~\citep{watkins1992q} and double estimator~\citep{hasselt2010double} are the most representative methods to estimate $\mu_{*}$. ME estimation: $\hat{\mu}_{*}^{M E}(S)=\max _{i} \hat{\mu}_{i}(S)=\max _{i} E\left\{S_{i}\right\} \approx \mu_{*}$. DE splits the set $S$ into $S^{A}=\left\{S_{1}^{A}, \ldots, S_{M}^{A}\right\}$ and $S^{B}=\left\{S_{1}^{B}, \ldots, S_{M}^{B}\right\}$. DE estimation: $\hat{\mu}_{*}^{D E}\left(S^{A}, S^{B}\right)=\hat{\mu}_{a^{*}}\left(S^{B}\right)=E\left[S_{a^{*}}^{B}\right] \approx \mu_{*}$, with $a^{*}=\argmax _{i} \hat{\mu}_{i}(S^{A})$.
\subsubsection{Bias}
We start with representing the main results about the bias
of Maximum Estimator (ME) and Double Estimator (DE) reported in~\citep{van2013estimating}. As for the direction of the bias, ME is positively biased, while DE is negatively biased. ME is bounded by:
$
\operatorname{Bias}\left(\hat{\mu}_{*}^{M E}\right) \leq \sqrt{\frac{M-1}{M} \sum_{i=1}^{M} \frac{\sigma_{i}^{2}}{\left|S_{i}\right|}} .
$
For the bound of DE,~\citep{van2013estimating} conjectures the following lower bound:
$
\operatorname{Bias}\left(\hat{\mu}_{*}^{D E}\right) \geq-\frac{1}{2}\left(\sqrt{\sum_{i=1}^{M} \frac{\sigma_{i}^{2}}{\left|S_{i}^{A}\right|}}+\sqrt{\sum_{i=1}^{M} \frac{\sigma_{i}^{2}}{\left|S_{i}^{B}\right|}}\right).
$
$M$ means the number of sample means, $\sigma_{i}$ means the variance of the $i_{th}$ sample mean. For the bias of DPAV estimator, we have the following bounds.
\paragraph{Theorem 2.} For any given set $X$ of $M$ random variables:
$
\operatorname{Bias}\left(\hat{\mu}_{*}^{DPAV}\right) \leq \operatorname{Bias}\left(\hat{\mu}_{*}^{M E}\right),
$
and
$
\operatorname{Bias}\left(\hat{\mu}_{*}^{DPAV}\right) \geq \operatorname{Bias}\left(\hat{\mu}_{*}^{D E}\right).
$

\paragraph{Explanation.} ME uses the maximum of sample means to estimate the ground truth maximal expected value (MEV), while DPAV takes the partial average over the maximum and minimum of sample means. The minimum will shift the DPAV estimation towards the ground truth, so the upper bound of the DPAV estimator bias will be smaller than that of ME. DE uses the minimum of sample means to estimate the ground truth in the worst case, however, the DPAV estimator mitigates this bias through importing the maximum into the partial average shifting the estimation towards the ground truth. So its lower bound is larger than that of DE.
\subsubsection{Variance}
Since the MSE loss of an estimator is the sum of its squared bias and its variance, so we should also consider its variance to evaluate its goodness.~\citet{van2013estimating} proved that both the variance of ME and the one of DE could be upper bounded by the sum of variances of sample means: 
$
\operatorname{Var}\left(\hat{\mu}_{*}^{M E/DE}\right) \leq \sum_{i=1}^{M} \frac{\sigma_{i}^{2}}{\left|S_{i}\right|}.
$
\paragraph{Theorem 3.} The variance of $DPAV$ estimator is upper bounded by:
$
\operatorname{Var}\left(\hat{\mu}_{*}^{DPAV}\right) \leq  \frac{\sigma_{Max/Min}^{2}}{\left|S_{Max/Min}\right|}\leq \sum_{i=1}^{M} \frac{\sigma_{i}^{2}}{\left|S_{i}\right|}.
$
\paragraph{Explanation.} Because the DPAV estimator utilizes the partial average between the maximum and minimum of sample means to estimate the ground truth. The weights assigned to the maximum and minimum are in the range (0,1), and the sum of weights is 1. According to the variance math properties~\citep{casella2021statistical}, the estimation variance is smaller than the larger one among the variances of maximum and minimum of sample means. Therefore, it is also smaller than the maximal variance of all sample means.

\section{Experiments}
\subsection{Dataset and Evaluation Metrics}
We evaluate the DPAV DQN method and baselines on three public task-completion dialogue datasets\footnote{\citep{zhao2021efficient} argued that the three tasks have been widely used in the research of dialogue policy.}: movie-ticket booking~\citep{li2016user,li2017end}, restaurant reservation and taxi ordering~\citep{li2018microsoft}.
The statistics of the datasets are given in Table \ref{table:statistics} (see Appendix~\ref{sec:appendixdata} for details)

The evaluation metrics are \textbf{success rate} and \textbf{averaged reward}. Success rate is the ratio of the number of tasks successfully completed by the dialogue system in evaluation to the total number of dialogues in the test set. Averaged reward refers to the average of the cumulative rewards obtained by the dialogue system for completing each dialogue of the test set.
\begin{table}[t]
\begin{tabular}{l}
\hline
~~~~~~Task~~~~~~~~~~~~~~~~~~~~~~  Intents Slots  User goals \\
\hline
Movie-ticket booking  ~\:~11  \:\:\:\:\:\:\:\:16  \:\:\:\:128 \\
Restaurant reservation ~\:11  \:\:\:\:\:\:\:\:30  \:\:\:\:3525\\
Taxi ordering  \,\:\:\:~~~~~~~\:~~~\:\:11  \:\:\:\:\:\:\:\:29  \:\:\:\:2830 \\
\hline
\end{tabular}
\caption{The statistics of the datasets.}
\label{table:statistics}
\end{table}
\subsection{Baselines}
To benchmark our method performance, we use different DQN variants as baselines in dialogue policy module for comparison: (1)~\textbf{DQN} policy is learned with standard DQN algorithm~\cite{mnih2015human}. (2)~\textbf{Duel DQN} policy is learned by the duel network structure~\citep{wang2016dueling}.(3) \textbf{Double DQN} policy uses Double Estimator of Q-learning to train~\citep{d2016estimating}. (4) \textbf{Averaged DQN} policy is trained by taking average over multiple action values of target networks~\citep{anschel2017averaged}. (5) \textbf{Maxmin DQN} policy uses the minimum of multiple maximums from different ensemble units to estimate the ground truth maximal action value in a selective process~\citep{lan2020maxmin}. (6) \textbf{SUNRISE} policy trains with weighted Bellman backups from multiple networks~\citep{lee2021sunrise}. Our model \modelname~ uses a value function instead of a combination of multiple value functions to tailor the maximum and the minimum action value.

We conduct two $\lambda$ searching schemes: neural network (NN) searching and heuristic searching. We also analyze the influence of different initial value $\lambda_{0}$ in the heuristic searching. So we have following models in the experiment: 
(1)~\textbf{Lambda}\textit{X} is the heuristic searching version of the DPAV DQN. The floating number X is the initial value $\lambda_{0}$ with the range $(0,1)$. And \textbf{Lambda}\textit{X} (e.g. Lambda0.5, Lambda0.6) searches different floating numbers \textit{X} for initial $\lambda_{0}$ in the heuristic searching.
(2)~\textbf{LambdaNet} is the neural network searching version of the DPAV DQN. It trains a NN to find a value for 
$\lambda_{t}$ for each dialogue state $s_{k}$ in the training process. Here, $\lambda_{t}$ means the value $\lambda$ in the training episode $t$, and $s_{k}$ represents the dialogue state $k$ sampled from the experience replay buffer of reinforcement learning.
\subsection{Implementation Details}
This work is implemented with PyTorch toolkit. Compared with the standard DQN algorithm, we change the loss with the one defined by DPAV DQN in Algorithm~\ref{alg:dpav}.
For these RL-based dialogue policies, action value network $Q(\cdot)$ is a MLP with one hidden layer of 80 hidden nodes. ReLU is the activation function. A greedy policy is used in the evaluation. All neural networks warm start 120 episodes using the same rule-based policy before training and are trained with the same hyper-parameters. We follow the default hyper-parameters of the user simulator setting. The discount factor $\gamma$ for future reward is 0.9. The batch size is 16, and the learning rate is 0.001. The test set size in the movie domain and other domains is set to 100 and 500, respectively. All baselines are based on DQN for a fair comparison. We set $L=40$ as the maximum of dialogue turns in all domains. The heuristically searched decay rate $d$ and decay interval of the DPAV estimator in the movie domain and other domains are set to (0.75, 15 train iterations) and (0.9965, 30 train iterations), respectively. For specific parameters of each model and the user simulator, we refer to Appendix~\ref{sec:appendiximple}.
\subsection{Main Results}
\begin{figure*}[ht!]
\centering
\begin{subfigure}[b]{.333\textwidth}
  \centering
  \includegraphics[width=\textwidth]{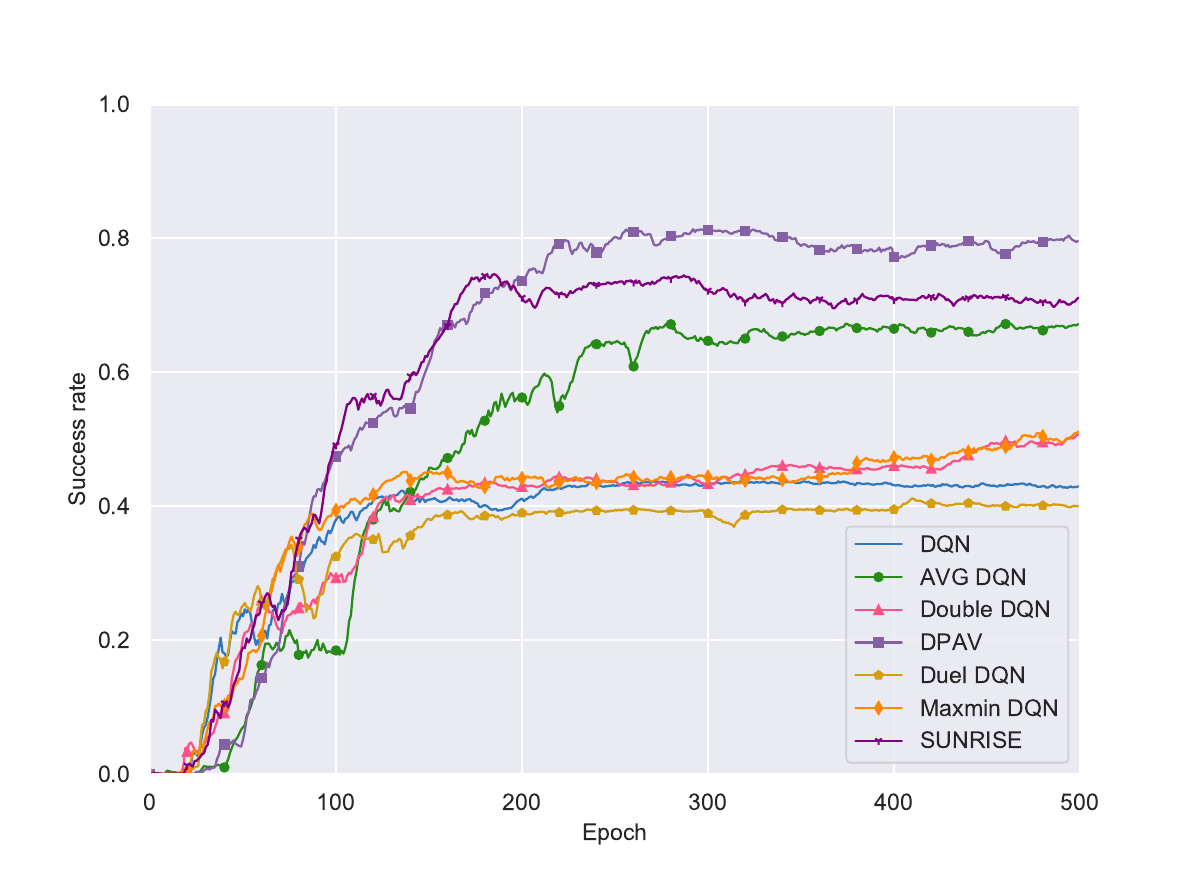}
  \caption{Movie domain success rate}
  \label{img:movie_policy}
\end{subfigure}%
\begin{subfigure}[b]{.33\textwidth}
  \centering
  \includegraphics[width=\textwidth]{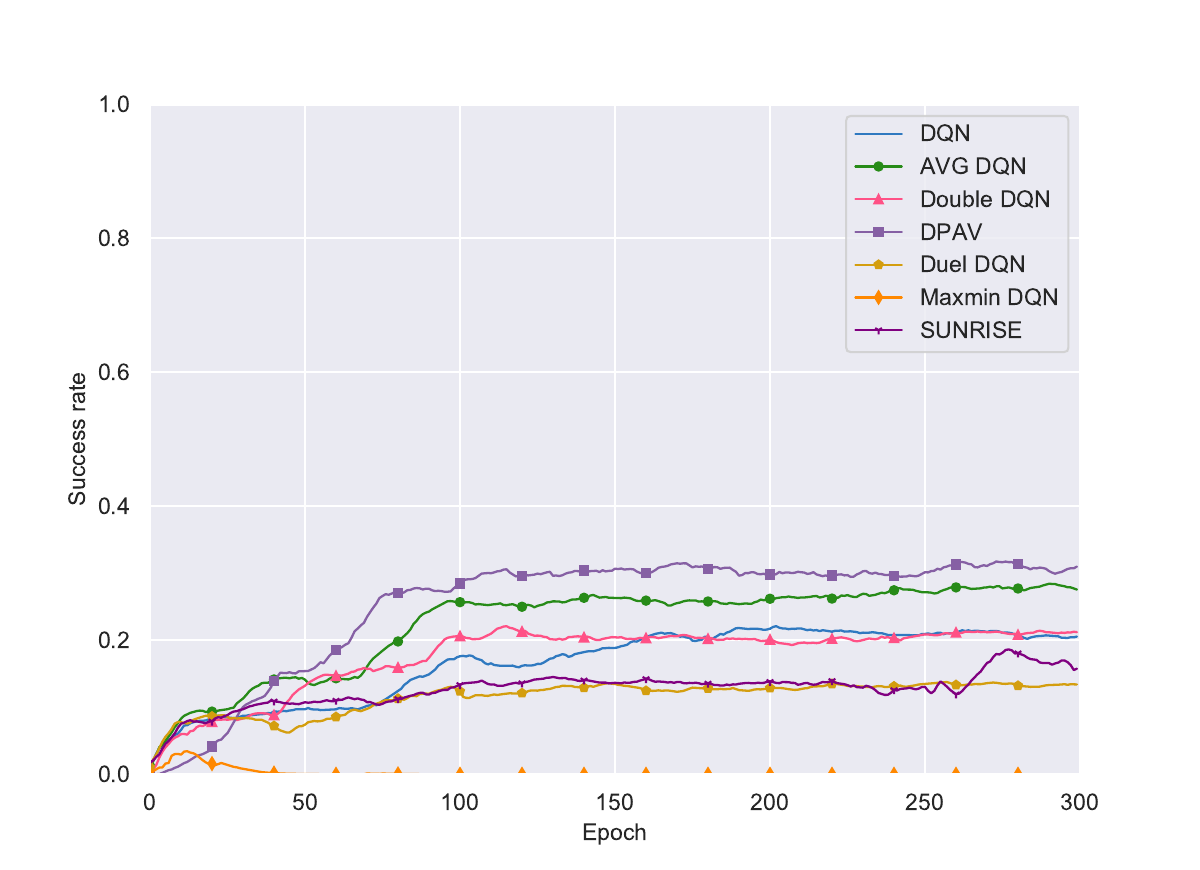}
  \caption{Restaurant domain success rate}
  \label{img:rest_policy}
\end{subfigure}%
\begin{subfigure}[b]{.33\textwidth}
  \centering
  \includegraphics[width=\textwidth]{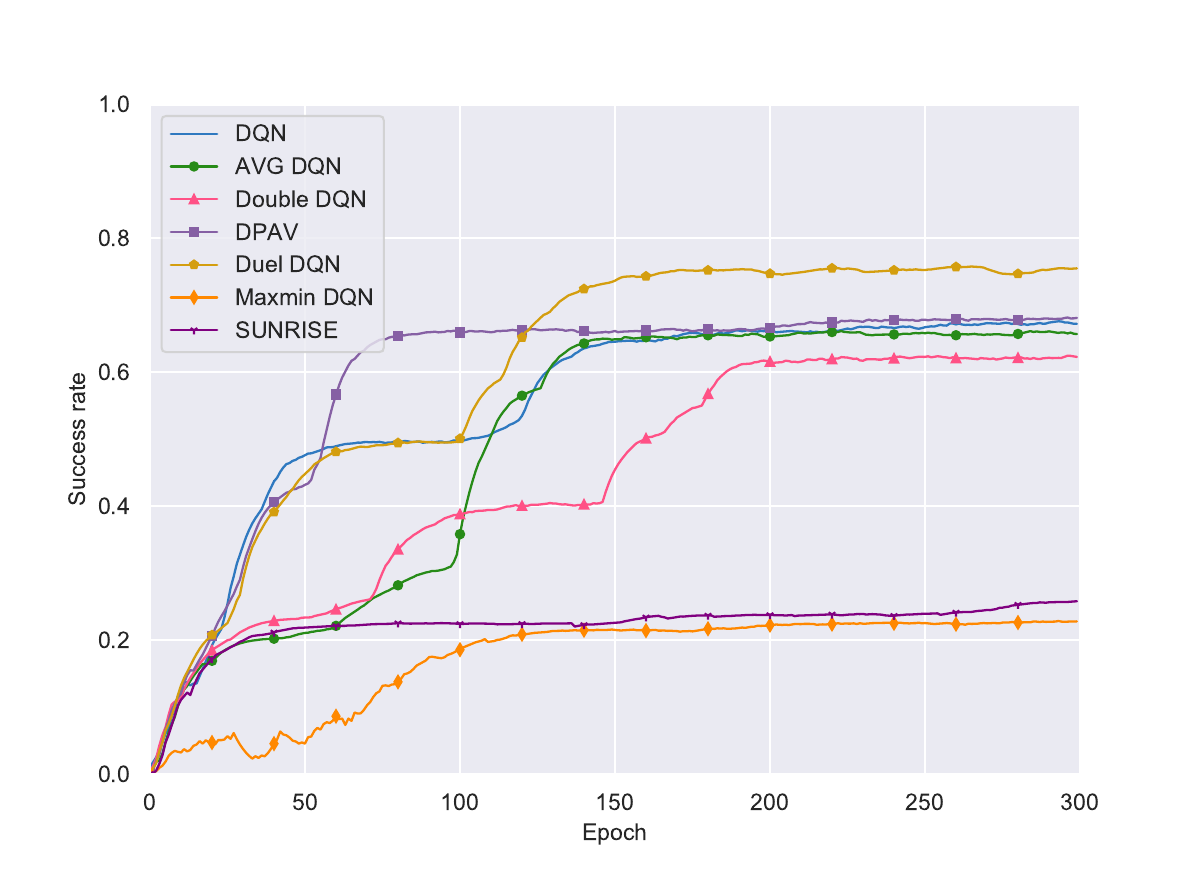}
  \caption{Taxi domain success rate}
  \label{img:taxi_policy}
\end{subfigure}
\begin{subfigure}[b]{.333\textwidth}
  \centering
  \includegraphics[width=\textwidth]{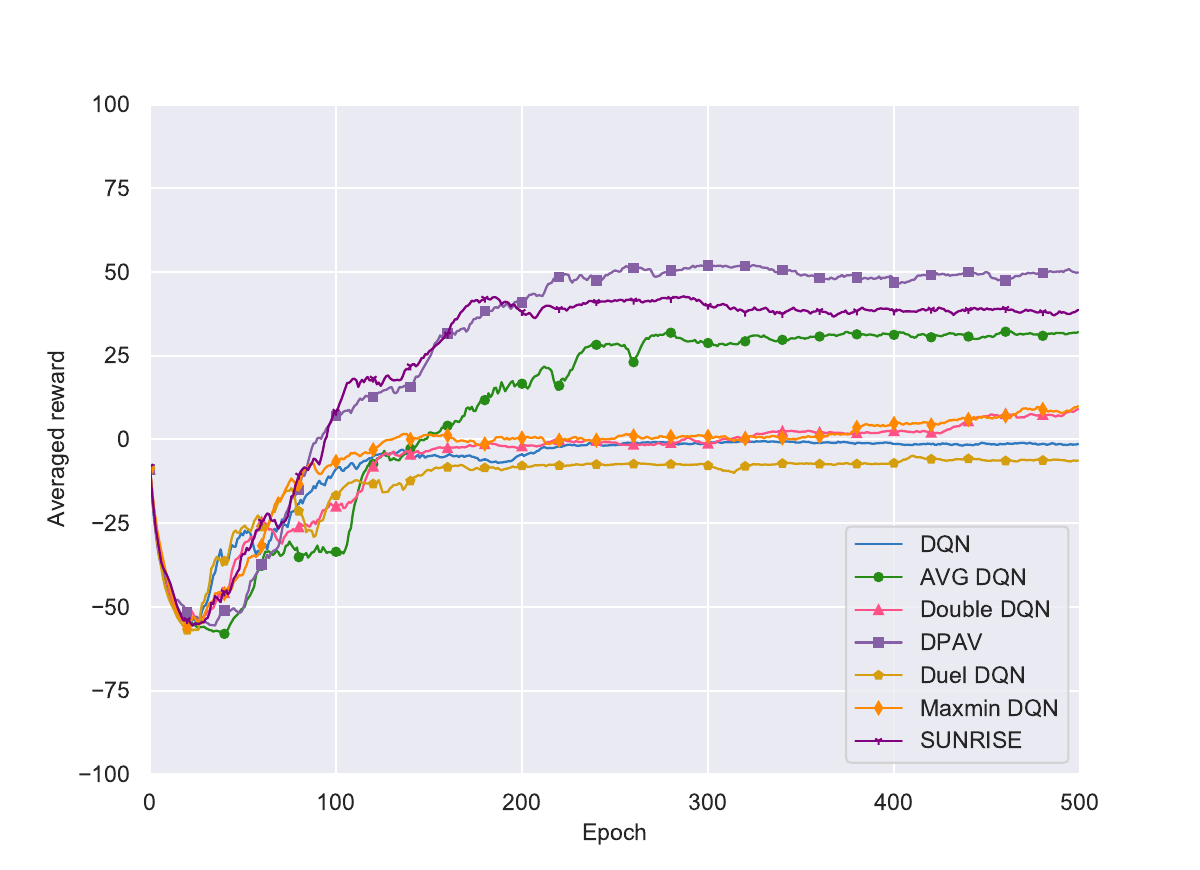}
  \caption{Movie domain average reward}
  \label{img:movie_reward}
\end{subfigure}%
\begin{subfigure}[b]{.33\textwidth}
  \centering
  \includegraphics[width=\textwidth]{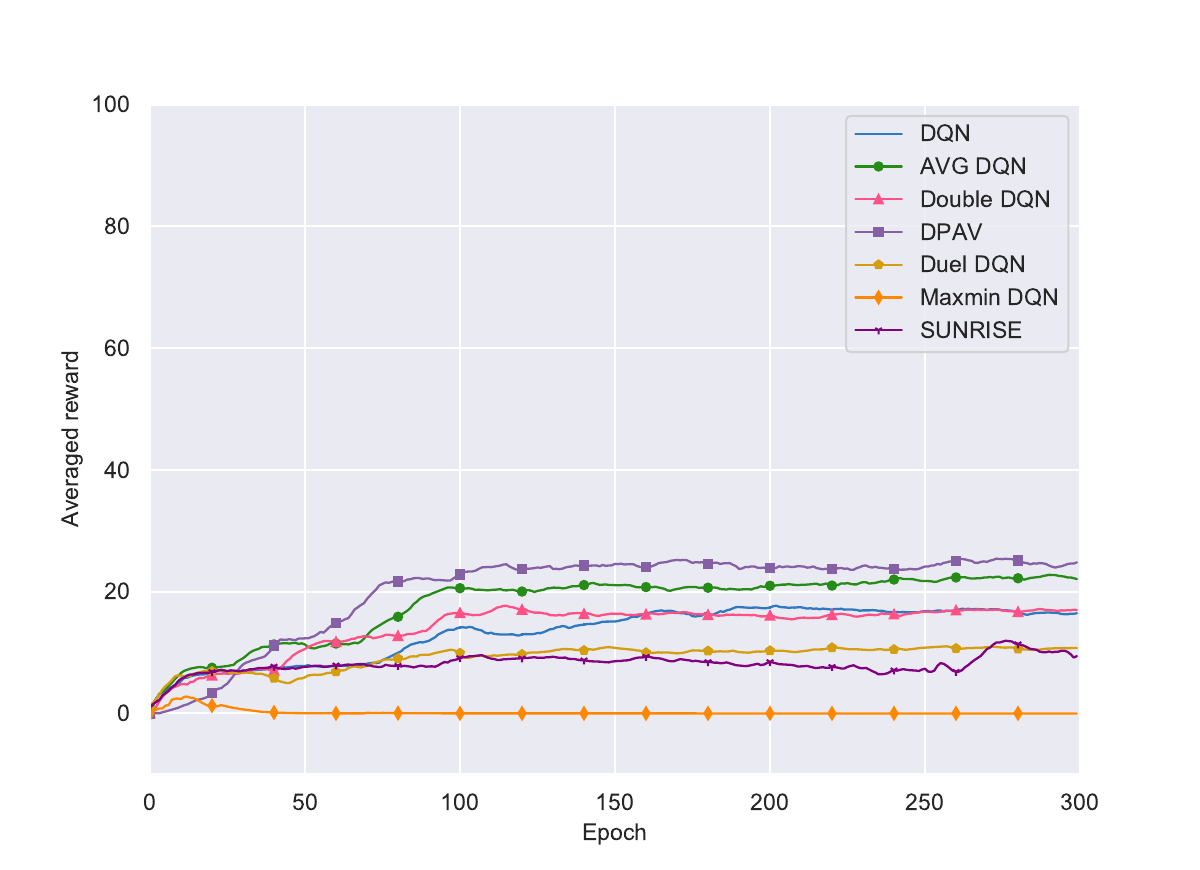}
  \caption{Restaurant domain average reward}
  \label{img:rest_reward}
\end{subfigure}%
\begin{subfigure}[b]{.33\textwidth}
  \centering
  \includegraphics[width=\textwidth]{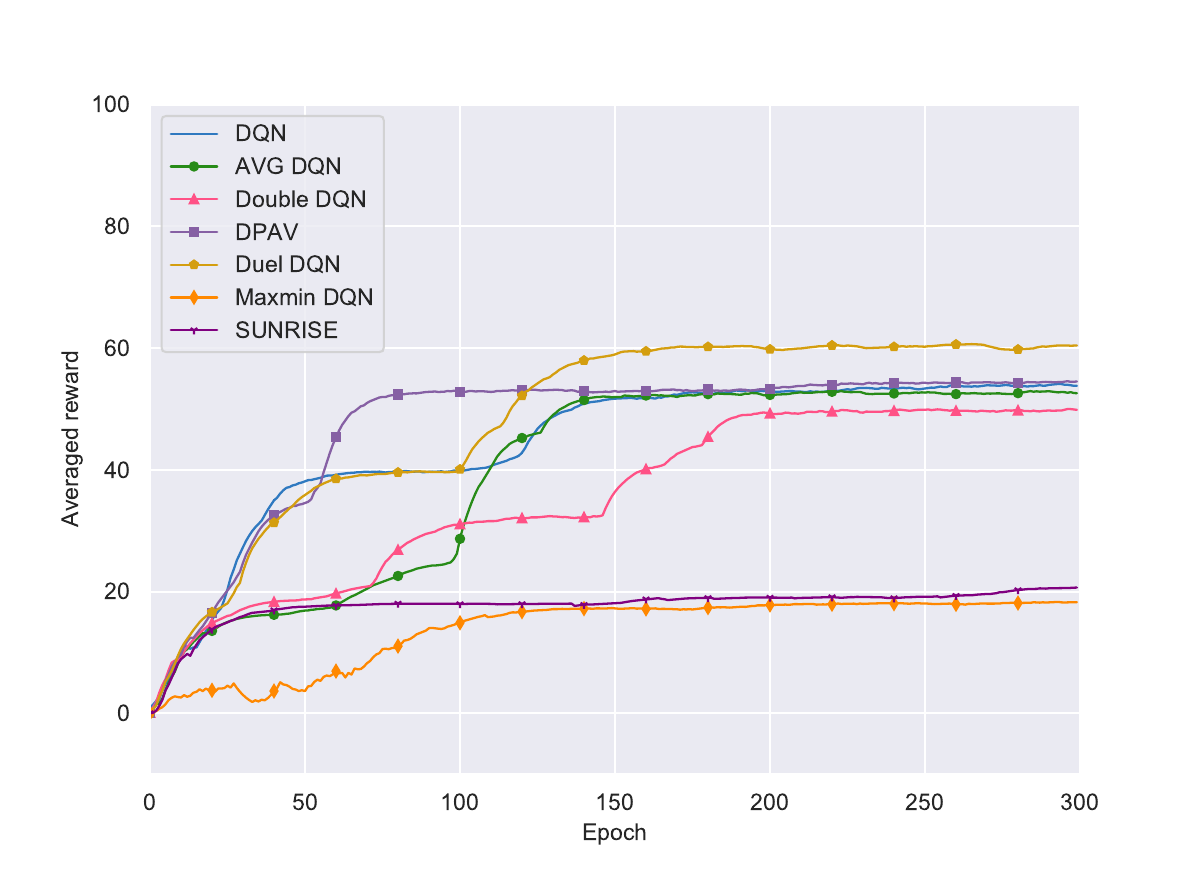}
  \caption{Taxi domain average reward}
  \label{img:taxi_reward}
\end{subfigure}
\begin{subfigure}[b]{.33\textwidth}
  \centering
  \includegraphics[width=\textwidth]{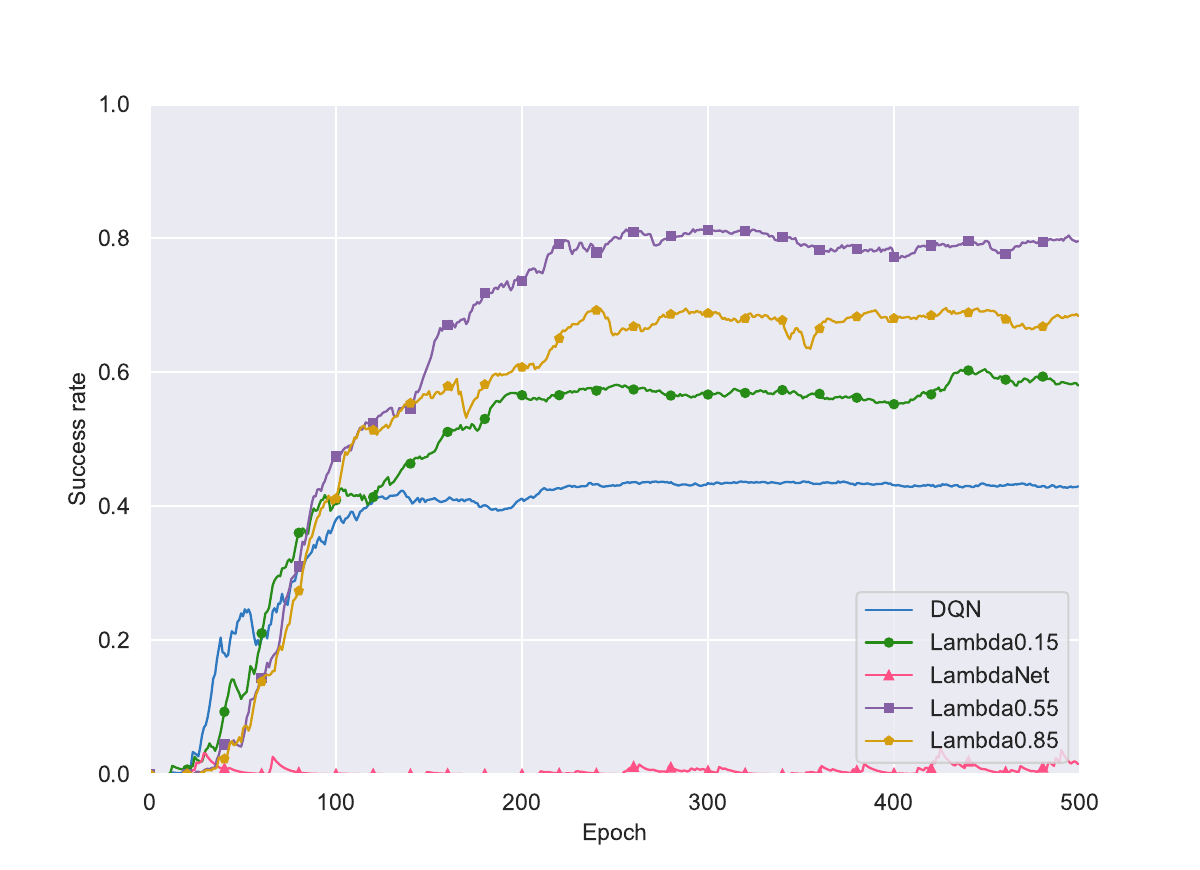}
  \caption{Movie domain parameter study}
  \label{img:movie_ablation}
\end{subfigure}%
\begin{subfigure}[b]{.33\textwidth}
  \centering
  \includegraphics[width=\textwidth]{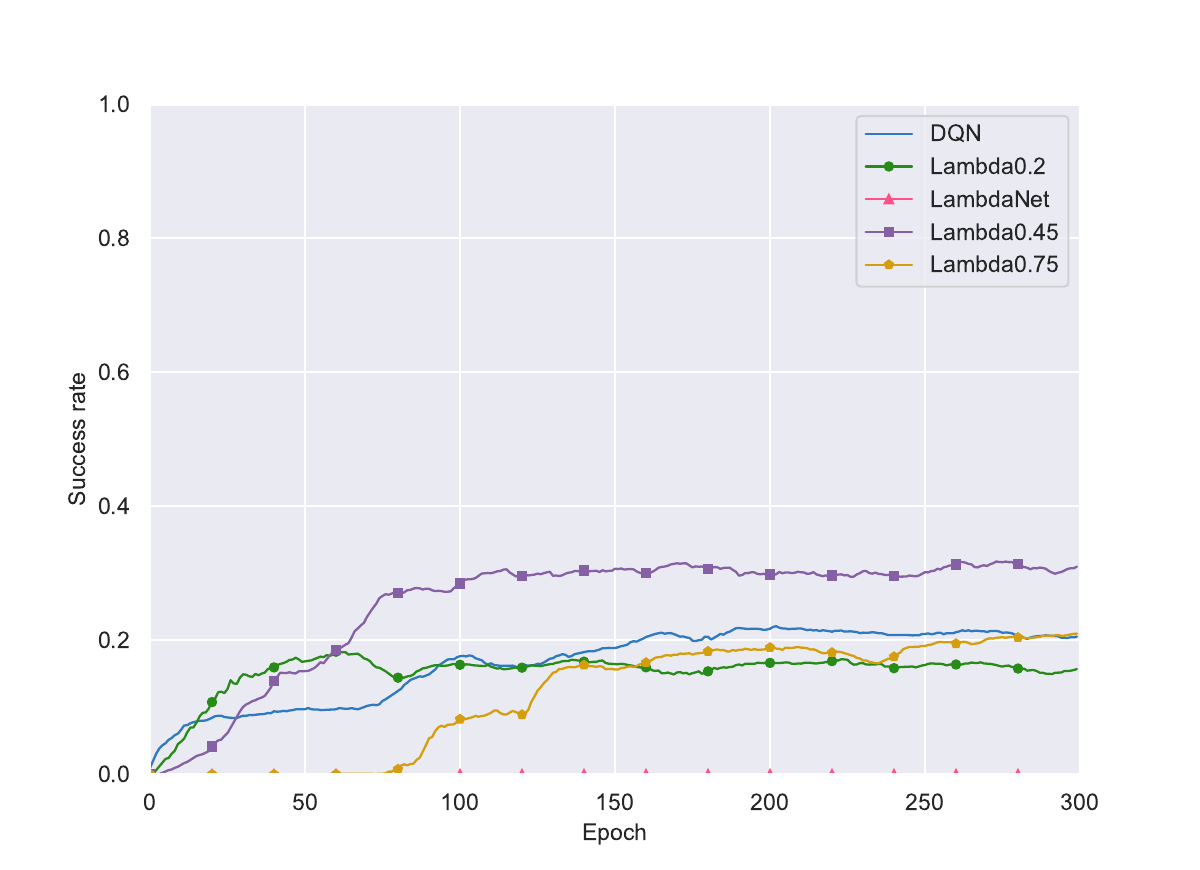}
  \caption{Restaurant domain parameter study}
  \label{img:rest_ablation}
\end{subfigure}%
\begin{subfigure}[b]{.33\textwidth}
  \centering
  \includegraphics[width=\textwidth]{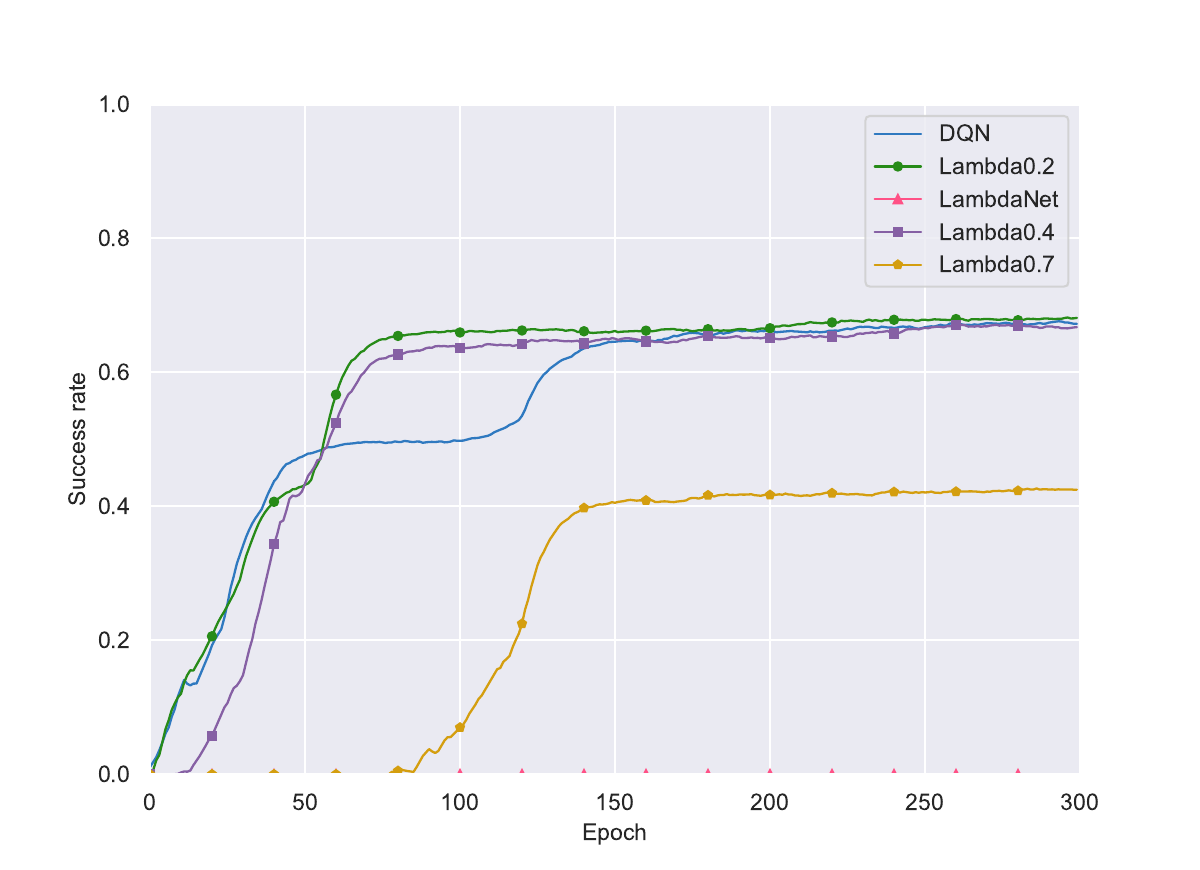}
  \caption{Taxi domain parameter study}
  \label{img:taxi_ablation}
\end{subfigure}
\caption{The \textbf{top} row shows the learning curves of dialogue policies. The X-axis is the number of training epochs and the Y-axis is the success rate of dialogue policies on the test dataset.  The \textbf{second} row shows the averaged reward of each dialogue in the test dataset. 
The \textbf{third} row shows the influences of different initial $\lambda$ values and value search schemes. The X-axis and Y-axis are the same as those of the top row. Each learning curve is averaged over 3 runs on the test dataset.}
\label{fig:all_tests}
\end{figure*}
\begin{figure}[ht!]
\includegraphics[width=0.5\textwidth]{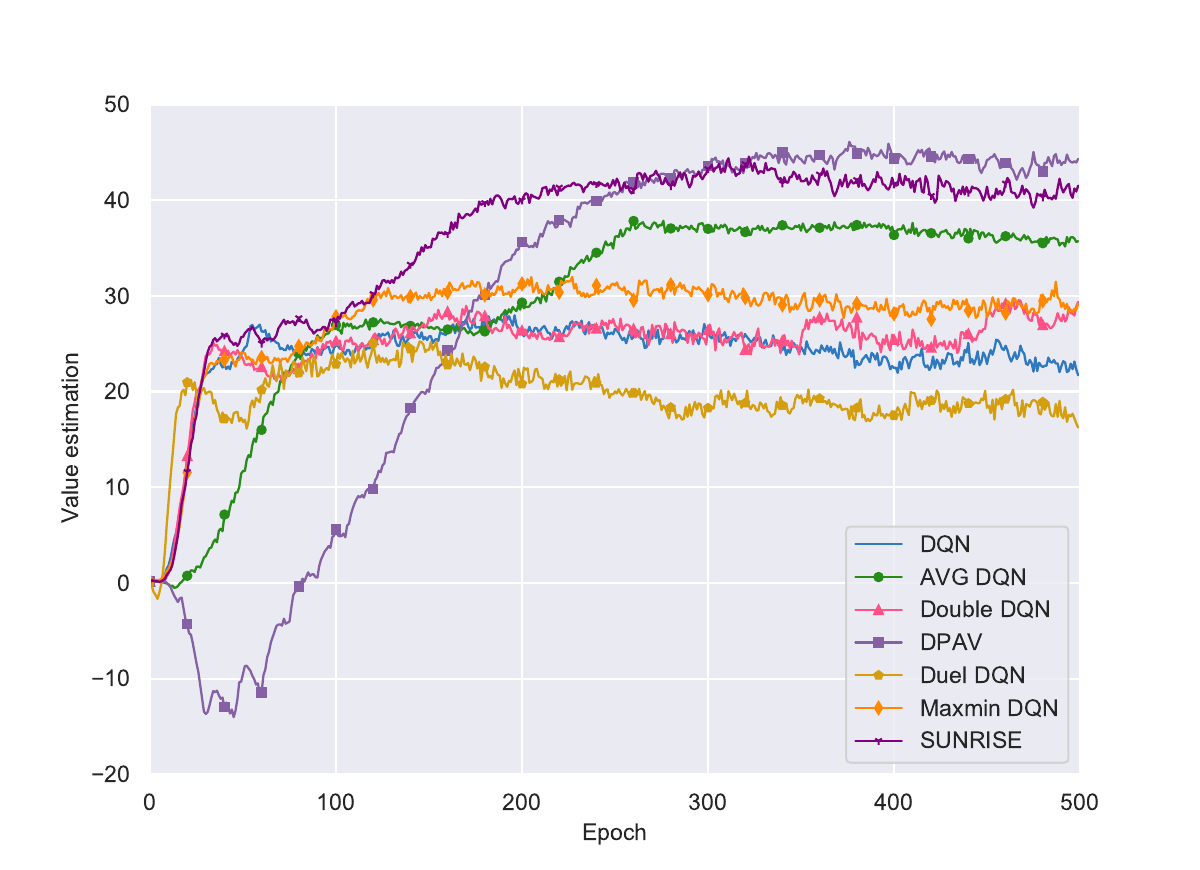}
\centering
\caption{The learning curves of the averaged maximal action value of the dialogue starting state when dialogue policies are evaluated on the movie test set during the training. The Y-axis means the averaged maximal action value of the starting state. }
\label{fig:mqmax}
\end{figure}
The main simulation results are reported in Figure~\ref{fig:all_tests},
we evaluate each dialogue policy performance in terms of success rate and averaged reward. The top two rows of Figure~\ref{fig:all_tests} show DPAV DQN consistently outperforms DQN. The overestimation error in target Q values gets propagated into the DQN Q values, while DPAV DQN reduces the overestimation error then its Q values will be less biased. So it 
more correctly creates dialogue utterances based on the Q values and achieves a better success rate and averaged reward. 

Our DPAV DQN method performs better than the baselines in terms of general performance. Since the training starts with the experience pool initialized by the the same rule-based dialogue policy, the models' performance in the very first few episodes is very similar. After that, the performance improved for all models, but much rapidly for DPAV DQN, which finally converges to a higher success rate and averaged reward. As we claimed above, the DPAV estimator reduces the overestimation error propagated into its model Q values and results in better action values estimation. Ensemble models performance relies on its number of networks. With a limited number of networks, as mentioned in the Related Work, Averaged DQN still suffers overestimation bias, Maxmin DQN has estimation bias from the coarse estimator and SUNRISE only down-weights the biased estimation.
For non-ensemble models, Duel DQN suffers overestimation with the Maximum Estimator, while Double DQN has underestimation bias~\citep{anschel2017averaged}. These drawbacks of the baselines get their biased loss propagated into the policy model Q values and hurt the accuracy of the policy models. So their performance (i.e., success rate and averaged reward) cannot improve further after reaching a certain level. The training efficiency and performance of DPAV DQN in comparison validate the effectiveness of our model. 

However, in the taxi domain,
Duel DQN outperforms other dialogue policies. DPAV DQN only slightly improves the results compared to DQN but it converges faster than DQN. Because sometimes there is no explicitly preferable action for a state, so the action values of the state will be similar~\citep{thrun1993issues}, and the DPAV estimator cannot notably reduce estimation bias through averaging between maximum and minimum action values. But the DPAV estimator still could estimate better than other baselines (except for Duel DQN) as shown in the results. Duel DQN uses the duel network structure to estimate action values~\citep{wang2016dueling}, which is helpful for recognizing the correct action when confronted with confusing states. 
\subsection{Influence of Parameter $\lambda$}
Intuitively, the optimal $\lambda$ should seek the best trade-off between the estimated maximum and minimum that could be used to train the dialogue policy properly. It is a non-trivial optimization problem because the distribution of action values $Q\left(s_{t}, \cdot\right)$ at state $s_{t}$ is constantly updated, and the optimal $\lambda$ for $s_{t}$ should be adjusted accordingly. The third row of Figure~\ref{fig:all_tests} shows that with the neural network (NN) searching almost each dialogue policy evaluation has zero success rate and can not converge. Since the distribution and the optimal $\lambda$ are changing for the same state $s_{t}$, the fixed $\lambda$ searched by the neural network does not work. This validates that the $\lambda$ for $s_{t}$ is dynamic, and a fixed $\lambda$ leads to bad performances.

It is a difficult problem to calculate the exact ground truth maximal action value, so existing works use estimators to approximate it~\citep{lan2020maxmin,lee2021sunrise,anschel2017averaged}. DPAV DQN uses the DPAV estimator for the approximation. In the heuristic searching version of DPAV DQN, $\lambda_{0}$ is a very important initial value for the DPAV estimator. $\lambda_{0}$ is the initial weight of the minimum, because finally we will give the total confidence to our model, the weight of the maximum will be nearly 1. So the lower bound for $\lambda_{0}$ is 0. Since model reduces the overestimation of the maximum through shifting the estimate towards the minimum action value. It is a trade-off, so the upper bound for $\lambda_{0}$ is 1. It is problem dependent and should be set in a range (0,1).

As shown in the third row of Figure~\ref{fig:all_tests}, among three dialogue datasets: movie, restaurant and taxi, we empirically find that the initial value $\lambda_{0}$ around 0.5 results in good performances. And other heuristic values degrade the dialogue policy performances. Since shifting the estimate towards the minimum action value too much or too less both causes the estimation bias of the ground truth maximal action value. These validate that $\lambda_{0}$ is problem dependent and $\lambda_{t}$ should decay to proper values to balance the maximum and the minimum along the training.
\subsection{Computational Complexity Comparison}
All baselines and DPAV DQN use various estimators or estimation tricks to approximate the ground truth maximal action value $Q_{*}(s_{t+1})$. Given the state $s_{t+1}$ as the input to all baselines and DPAV DQN, the time complexity of the forward propagation of every model unit has a similar time complexity besides the minor differences (e.g., addition). In order to facilitate the comparison, we denote the time complexity for the forward propagation of every model unit as O(N), here, N means the dimension of the vector input for forward propagation. In this comparison, we suppose each ensemble model has K model units. 

Combining the results of Figure~\ref{fig:all_tests} and Table~\ref{tab:comple}, DPAV DQN achieves better or comparable performances with a lower time complexity. Although the time complexity of ensemble models can be reduced by parallel computing, but that increases the space complexity. So, the overall computational complexity is still high and resource consuming.
\begin{table}
\centering
\begin{tabular}{lc}
\hline
\textbf{Model Name} & \textbf{Time Complexity}\\
\hline
\verb|DQN| & {O(N)} \\
\verb|Double DQN| & {O(N)} \\
\verb|Duel DQN| & {O(N)} \\ 
\verb|Averaged DQN| & {O(K*N)} \\ 
\verb|Maxmin DQN| & {O(K*N)} \\
\verb|SUNRISE| & {O(K*N)}  \\ 
\verb|DPAV DQN| & {O(N)}  \\\hline
\end{tabular}
\caption{Time complexity comparison among baselines and DPAV DQN. N refers to the dimension of the vector input for the forward propagation. K means the number of ensemble units. O measures the time complexity of models.}
\label{tab:comple}
\end{table}

\subsection{Results on Maximum Action Value}
In reinforcement learning, the action value $Q$ is the expectation of return $R_{t}$ that is the sum of the discounted rewards:
$Q(s, a)=E\left\{R_{t} \mid s_{t}=s, a_{t}=a\right\}$
$=E\left\{\sum_{k=0}^{\infty} \gamma^{k} r_{t+k+1} \mid s_{t}=s, a_{t}=a\right\}$.
Figure~\ref{fig:mqmax} shows learning curves\footnote{To save space, we only present the results on the movie dataset, and the results on other datasets are similar.} of the averaged maximum action value for the starting state on the test set, the value in dialogue context means how much return the dialogue policy assumes it could maximally receive from the starting state.

At the first few training epochs, we notice that the averaged maximum value of DPAV DQN is negative which is consistent with the averaged reward of its evaluation shown in Figure~\ref{img:movie_reward}, because at the early training stage, the policy quality is too low to finish most of the dialogues so the averaged reward is low and the averaged maximum action value should be low if the model Q values are accurate. But the values of other models are not consistent with and larger than the real averaged reward. Because the estimation bias of the loss makes that these models have inaccurate Q values, the maximum action value of these models is larger than the ground truth. 

The policy training based on these inaccurate Q values will be negatively affected. Only using Maximum Estimator (ME) will cause overestimation bias and even lead to worse policy quality, it can be observed from the curves of DQN and Duel DQN in Figure~\ref{fig:mqmax}. Averaged DQN and Maxmin DQN use ME in their single unit so the bias leads their Q functions to converge into inaccurate values, which prevents averaged maximum action values from improving further. SUNRISE down-weights the biased estimation and it is trained in such a way 
so that SUNRISE dialogue policy receives more rewards during the evaluation~\ref{img:movie_reward}. As shown in the Figure~\ref{fig:mqmax}, the averaged maximal action value of DPAV DQN remains the highest among the three datasets because its model gets trained better with the less biased loss and receives more return from successful dialogues during evaluation. This also coincides with the averaged reward from test dialogues in Figure~\ref{img:movie_reward}. This empirically validates that DPAV is a better estimator than others because of less estimation bias.
\section{Conclusion}
This paper is the first to investigate the negative effects of the overestimation problem in task-completion dialogue systems. We propose the DPAV estimator to mitigate this problem of Q-learning. We also theoretically prove convergence and derive the upper and lower bounds of the estimation bias compared with those of other methods. The resulting 
DPAV DQN model 
is empirically evaluated on three dialogue datasets and 
achieves better or comparable results with lower computational load compared to state-of-the-art baselines.
\section*{Acknowledgements}
We would like to thank Xiumei Zhao, Dan Li, Wentao Huang and Pengjie Ren for reading the paper draft. This research was partially supported by the China Scholarship Council.  All content represents the opinion of the authors, which is not necessarily shared or endorsed by their respective employers and/or sponsors. Marie-Francine Moens is supported by the ERC Advanced Grant CALCULUS (788506).
\bibliography{acl_latex}
\bibliographystyle{acl_natbib}

\appendix
\section{Appendix}
\subsection{Lemma}
\label{sec:appendixlemma}
\textbf{Lemma 1}~\citep{hasselt2010double}. Let $\left(\beta_{t}, \Delta_{t}, F_{t}\right)$ be a stochastic process, where $\beta_{t}, \Delta_{t}, F_{t}: X \mapsto \mathbb{R}$ satisfy,
$$
\Delta_{t+1}\left(x_{t}\right)=\left(1-\beta_{t}\left(x_{t}\right)\right) \Delta_{t}\left(x_{t}\right)+\beta_{t}\left(x_{t}\right) F_{t}\left(x_{t}\right)
$$
where $x_{t} \in X$ and $t=0,1,2, \ldots$. Let $P_{t}$ be a sequence of increasing $\sigma$-fields such that $\beta_{0}$ and $\Delta_{0}$ are $P_{0^{-}}$ measurable and $\beta_{t}, \Delta_{t}$, and $F_{t-1}$ are $P_{t}$-measurable, with $t \geq 1$. Assume that the following conditions are satisfied:\\
1. The set $X$ is finite $($ i.e. $|X|<\infty)$.\\
2. $\beta_{t}\left(x_{t}\right) \in[0,1], \sum_{t} \beta_{t}\left(x_{t}\right)=\infty, \sum_{t} \beta_{t}^{2}\left(x_{t}\right)<\infty$ w.p. 1 , and $\forall x \neq x_{t}: \beta_{t}(x)=0$. $\beta_{t}\left(x_{t}\right)$ is the step size of the update.\\
3. $\left\|\mathbb{E}\left\{F_{t} \mid P_{t}\right\}\right\| \leq \kappa\left\|\Delta_{t}\right\|+c_{t}$, where $\kappa \in[0,1)$ and $c_{t} \rightarrow 0$ w.p.1.\\
4. $\mathbb{V}\left\{F_{t}\left(x_{t}\right) \mid P_{t}\right\} \leq C\left(1+\kappa\left\|\Delta_{t}\right\|\right)^{2}$, where $C$ is some constant.\\
where $\mathbb{V}\{\cdot\}$ denotes the variance and $\|\cdot\|$ denotes the maximum norm. Then $\Delta_{t}$ converges to zero with probability one.\\
\\
Proof. See~\citep{singh2000convergence}.
\subsection{DPAV Q-learning Convergence Proof}
\label{sec:appendixdpavpr}
\paragraph{Proof.} We apply Lemma 1 with $X=\mathcal{S} \times \mathcal{A}$, $\Delta_{t}=Q_{t}(s_{t}, a_{t})-q_{*}(s_{t}, a_{t})$, $\beta_{t}=\alpha_{t}$, $\beta_{t}$ is also the step size,  $P_{t}=\left\{Q_{0}, s_{0}, a_{0}, \alpha_{0}, r_{1}, s_{1}, \ldots, s_{t}, a_{t}\right\}$ and
\begin{equation}
F_{t}\left(s_{t}, a_{t}\right)=r_{t+1}+\gamma Q_{d p a v}\left(s_{t+1}, \cdot\right)-q_{*}\left(s_{t}, a_{t}\right),
\end{equation}
where
\begin{equation}
\begin{aligned}
Q_{d p a v}\left(s_{t+1}, \cdot\right) &=\left(1-\lambda_{t}\right) Q_{t}\left(s_{t+1}, a_{\max }\right) \\
&+\lambda_{t} Q_{t}\left(s_{t+1}, a_{\min }\right).
\end{aligned}
\end{equation}
And $a_{max}=\arg \max _{a^{\prime}} Q_{t}\left(s_{t+1}, a^{\prime}\right)$ while $a_{min}=\arg \min _{a^{\prime\prime}} Q_{t}\left(s_{t+1}, a^{\prime\prime}\right)$.
The first condition of the Lemma 1 is satisfied because $|\mathcal{S} \times \mathcal{A}|<\infty$. The second condition of Lemma 1 is met by the third condition of Theorem 1. Because the absolute value of reward $|r|<\infty \Longrightarrow \forall t:$ $\mathbb{V}\left\{r_{t+1} \mid P_{t}\right\}<\infty$. Since $Q_{t}$ is the expected cumulative reward in Q-learning and $F_{t}\left(s_{t}, a_{t}\right)$ is composed of reward $r$, so $\forall t: \mathbb{V}\left\{r_{t+1} \mid P_{t}\right\}<\infty \Longrightarrow \forall t: \mathbb{V}\left\{F_{t}\left(s_{t}, a_{t}\right) \mid P_{t}\right\}<\infty$, the fourth condition of the Lemma 1 is sufficed. This leaves to show that the third condition of the Lemma 1 on the expected contraction of $F_{t}$ holds. We can write
$$
\begin{aligned}
\scriptsize
&F_{t}\left(s_{t}, a_{t}\right)=r_{t+1}+\gamma(\left(1-\lambda_{t}\right) Q_{t}\left(s_{t+1}, a_{\max }\right) \\
&+\lambda_{t} Q_{t}\left(s_{t+1}, a_{\min }\right))-q_{*}\left(s_{t}, a_{t}\right)\\
&=r_{t+1}+\gamma(Q_{t}\left(s_{t+1}, a_{\max }\right)-\lambda_{t}Q_{t}\left(s_{t+1}, a_{\max }\right)\\
&+\lambda_{t} Q_{t}\left(s_{t+1}, a_{\min }\right))-q_{*}\left(s_{t}, a_{t}\right)\\
&=r_{t+1}+\gamma Q_{t}\left(s_{t+1}, a_{\max }\right)-q_{*}\left(s_{t}, a_{t}\right)\\
&+\gamma\lambda_{t}\left(Q_{t}\left(s_{t+1}, a_{\min }\right)-Q_{t}\left(s_{t+1}, a_{\max}\right)\right)\\
&=r_{t+1}+\gamma Q_{t}\left(s_{t+1}, a_{\max }\right)-q_{*}\left(s_{t}, a_{t}\right)\\
&+\gamma\lambda_{t}Q_{sub}\\
&=F_{t}^{\prime}\left(s_{t}, a_{t}\right)+\gamma\lambda_{t}Q_{sub},
\end{aligned}
$$
where $F_{t}^{\prime}$ is the value of $F_{t}$ if normal Q-learning would be under consideration, and $Q_{sub}=Q_{t}\left(s_{t+1}, a_{\min }\right)-Q_{t}\left(s_{t+1}, a_{\max}\right)$. Since it is well known that $\forall t$ : $\left\|\mathbb{E}\left\{F_{t}^{\prime} \mid P_{t}\right\}\right\| \leq \gamma\left\|\Delta_{t}\right\|$~\citep{hasselt2010double}, it follows that,
$
\begin{aligned}
\scriptsize
&\left\|\mathbb{E}\left\{F_{t} \mid P_{t}\right\}\right\|\\ &=\left\|\mathbb{E}\left\{F_{t}^{\prime} \mid P_{t}\right\}\right\|+\gamma \lambda_{t}\left\|\mathbb{E}\left\{Q_{sub} \mid P_{t}\right\}\right\| \\
& \leq \gamma\left\|\Delta_{t}\right\|+\gamma \lambda_{t}\left\|\mathbb{E}\left\{Q_{sub} \mid P_{t}\right\}\right\|
\end{aligned}
$
Since in DPAV Q-learning, the $\lambda_{t}$ will decay as $\lambda_{t+1}= \lambda_{t} * d$. When $t \rightarrow \infty$, given $\varepsilon>0, \exists t_{0}: \forall t \geq t_{0} \Longrightarrow \lambda_{t}<\varepsilon \Longrightarrow$ $\lim _{t \rightarrow \infty} \lambda_{t}=0$. Therefore, it suffices to show that $c_{t}=$ $\gamma \lambda_{t}Q_{sub} \rightarrow 0$ w.p.1. Since all the conditions of lemma 1 are satisfied, it holds that, $\forall s, a: Q_{t}(s, a) \rightarrow q_{*}(s, a)$ w.p.1.

\section{Appendix}
\label{sec:appendixalgorithm}
\subsection{Dataset details}
\label{sec:appendixdata}
Table~\ref{table:statistics} lists the number of intents, slots and users goals in the three datasets used in the evaluation. And Table~\ref{table:annotated} shows all annotated dialogue acts and slots in details. 
Task-oriented dialogue systems are designed to help users to complete a specific goal \textit{G}. Even though the dialogue system knows nothing about the user goal explicitly, the whole dialogue progresses around this user goal \textit{G} implicitly. In order to explain the user goal better, we take a user goal as an example from the movie domain:
\begin{equation}
\begin{gathered}
\mathbf{G o a l}=\left(C=\left[\begin{array}{l}
\text { moviename }=\text { Enter } \\
\text { the Dragon } \\
\text { actor }=\text { BruceLee } \\
\text { date }=\text { today }
\end{array}\right],\right. \\
\left.R=\left[\begin{array}{l}
\text { theater }= \\
\text { starttime }=
\end{array}\right]\right).
\end{gathered}
\end{equation}
In this user goal, a user inquires the dialogue system about the theater and start time of a today’s movie
about the Enter the Dragon by Bruce Lee. The user goals are generated from the annotated datasets mentioned in Table~\ref{table:annotated}. The user goals extracted from the same dataset are then aggregated into a user goal set for that task. The user goals extracted from the same dataset are then aggregated into a user goal set for that task. When running a dialogue, the user simulator~\citep{li2016user} randomly samples a user goal from the user goal set to converse with the dialogue system. Helping the user to achieve specific user goals is the task to complete for dialogue systems. 
In this paper, we use the success rate and averaged reward as our main evaluation criteria. We do not use averaged turns into our criteria because overestimation bias mainly prevents the dialogue system from completing a task in a dialogue. This is explicitly with success rate and averaged reward, and this is not directly related with averaged turns. If and only if the dialogue system recognizes all constraints provided by users and informs all information that users want, and finally books the desired tickets successfully, the user goal is viewed as successful, and the dialogue policy received positive reward for success. The averaged reward means the averaged cumulative discounted reward received by dialogue system per dialogue.
\subsection{Implementation details}
\label{sec:appendiximple}
The size of the experience replay pool in the movie domain and other domains is set to 8000 and 10000, respectively.
The number of target networks in Averaged DQN, Maxmin DQN and SUNRISE is set to 4. The temperature parameter of SUNRISE is set to 2. The target network update period for Averaged DQN is set to 4. In the experiment, we use a user simulator to interact with dialogue systems. In the movie domain, the dialogue system receives a 2L reward if the dialogue finishes successfully and receives -L if it fails. Also, a fixed reward (-1) is given to the dialogue system for each dialogue turn. In the restaurant and taxi domains, the dialogue system receives a 2L reward if the dialogue finishes successfully and receives 0 if it fails. Also, a fixed reward (0) is given to the dialogue system for each dialogue turn. Under this setup, the dialogue datasets for experiments have varieties~\citep{li2016user}.

\begin{table*}[b]
    \centering
    \begin{tabular}{|l|l|l|l|l|}
    \hline
    Task       & Intents                                                                                                                                                                      & Slots                                                                                                                                                                                                                                                                                                                                                                                                         & Dialogues   \\ \hline
    Movie      & \begin{tabular}[c]{@{}l@{}}request,inform,\\ confirm\_question,\\ confirm\_answer,\\ greeting,closing,\\ deny,not\_sure,\\ multiple\_choice,\\ thanks,welcome\end{tabular}   & \begin{tabular}[c]{@{}l@{}}city,closing,\\ data,greeting,\\ distanceconstraints,\\ moviename,price,\\ numberofpeople,\\ starttime,state,\\ taskcomplete,theater,\\ teater\_chain,ticket,\\ video\_format,zip\end{tabular}                                                                                                                                                                                     & 280        \\ \hline
    Restaurant & \begin{tabular}[c]{@{}l@{}}request,inform,\\ confirm\_question,\\ confirm\_answer,\\ greeting,closing,\\ deny,not\_sure,\\ multiple\_choice,\\ thanks,welcome,\end{tabular}  & \begin{tabular}[c]{@{}l@{}}address,atmosphere,\\ choice,city,closing,\\ cuisine,date,food,\\ dress\_code,greeting,\\ distanceconstraints,\\ numberofkids,mealtype,\\ numberofpeople,\\ other,personfullname,\\ phonenumber,pricing,\\ rating,restaurantname,\\ restauranttype,seating,\\ starttime,state,zip,\\ result,occasion,\\ taskcomplete,reservation\end{tabular}                                      & 4103        \\ \hline
    Taxi       & \begin{tabular}[c]{@{}l@{}}request\_inform,\\ comfirm\_question,\\ confirm\_answer,\\ greeting,closing,\\ deny,not\_sure,\\ multiple\_choice,\\ thanks,welcome,\end{tabular} & \begin{tabular}[c]{@{}l@{}}car\_type,city,speed,\\ closing,car\_level,date,\\ distanceconstraints,\\ dropoff\_location,\\ zip,result,numberofkids,\\ greeting,name,driver\_id,\\ numberofpeople,other,\\ pickup\_location,state,\\ dropoff\_location\_city,\\ pickup\_location\_city,\\ pickup\_time,cost,\\ taxi\_company,mc\_list,\\ taskcomplete,taxi,budget,\\ emergency degree,drive\_level\end{tabular} & 3094       \\ \hline
    \end{tabular}
    \caption{The details of the datasets.~\citep{li2016user,li2018microsoft}}
    \label{table:annotated}
\end{table*}

\end{document}